\documentclass{article}

\usepackage{amsmath}
\usepackage[final]{neurips_2022}

\usepackage[utf8]{inputenc} % allow utf-8 input
\usepackage[T1]{fontenc}    % use 8-bit T1 fonts
\usepackage{hyperref}       % hyperlinks
\usepackage{url}            % simple URL typesetting
\usepackage{booktabs}       % professional-quality tables
\usepackage{amsfonts}       % blackboard math symbols
\usepackage{nicefrac}       % compact symbols for 1/2, etc.
\usepackage{microtype}      % microtypography
\usepackage{xcolor}         % colors
\usepackage{caption}
\usepackage{subcaption}
\usepackage{graphicx}

\title{Pitfalls of Conditional Batch Normalization for Contextual Multi-Modal Learning}

\author{Ivaxi Sheth \thanks{Corresponding author: ivaxi-miteshkumar-sheth.1@ens.etsmtl.ca}\\
  Mila, ÉTS Montréal\\
  \And
  Aamer Abdul Rahman \\
  Mila, ÉTS Montréal \\ 
  \And
  Mohammad Havaei \\
  ICH\\ 
  \AND
  Samira Ebrahimi Kahou \\
  Mila, ÉTS Montréal, CIFAR AI Chair \\
}
\usepackage{float}
\usepackage{wrapfig}
\usepackage{graphicx}
\begin{document}

\maketitle

\begin{abstract}
Humans have perfected the art of learning from multiple modalities through sensory organs. Despite their impressive predictive performance on a single modality, neural networks cannot reach human level accuracy with respect to multiple modalities. This is a particularly challenging task due to variations in the structure of respective modalities. Conditional Batch Normalization (CBN) is a popular method that was proposed to learn contextual features to aid deep learning tasks. This technique uses auxiliary data to improve representational power by learning affine transformations for convolutional neural networks. Despite the boost in performance observed by using CBN layers, our work reveals that the visual features learned by introducing auxiliary data via CBN deteriorates. We perform comprehensive experiments to evaluate the brittleness of CBN networks to various datasets, suggesting that learning from visual features alone could often be superior for generalization. We evaluate CBN models on natural images for bird classification and histology images for cancer type classification. We observe that the CBN network learns close to no visual features on the bird classification dataset and partial visual features on the histology dataset. Our extensive experiments reveal that CBN may encourage shortcut learning between the auxiliary data and labels. 
\end{abstract}

\section{Introduction}
Learning about complex relationships between real-world variables has been the key to progression of human kind \cite{goertzel2012architecture}. While some of these relationships are now intuitive to us, some are more complex mathematical relationships. With advancement of deep learning in the field of computer vision~\cite{he2016deep}, natural language processing \cite{vaswani2017transformer}, and speech recognition~\cite{bahdanau2016end}, learning complex relationships across modalities is still an open research question. Most of the current state-of-the-art algorithms, learn from single modality~\cite{dosovitskiy2020image,raffel2020exploring} which might limit the representation learnt by the model.

Processing data from multiple modalities can provide richer context to the respective modality\cite{wang2020learning}. For examples, video action recognition \cite{carreira2017quo, feichtenhofer2019slowfast, DBLP:journals/corr/abs-2104-13051} can be benefited by learning from the audio along with the visual features\cite{shi2021multi}. Such factors motivate \textit{context-based learning}, where auxiliary features are used to enhance the primary feature representation. To learn such contextual information, designing of methods that can learn from each individual modality along with their fusion is critical but a challenging task. In computer vision, Conditional Batch Normalization (CBN) based models that perform feature-wise affine transformations have shown to achieve strong visual reasoning\cite{perez2018film}.

CBN\cite{de2017modulating} relies on modulating the visual feature stream with auxiliary features in the intermediary layers of a CNN. The normalized activations produced by auxiliary network has been considered as a popular method to compute contextual information. Recent works have shown unreliability of batch normalization during distributional shifts, adversarial attacks and dependency on the dataset\cite{huang2022delving,galloway2019batch}. Lack of robustness of BN poses an immediate challenge for its conditional version, CBN. With adoption of CBN in various applications\cite{abdelnour2018clear,brock2018large,perez2018film}, it is important to study the quality of visual features learnt in presence of auxiliary data. 

In this work, we empirically and visually show that while CBN network which learns auxiliary data may surpass the performance of standard Batch Normalization(BN) network, they often fail to learn visual features altogether or partially. We devise a set of experiments to perform surgery into the CBN network. We observe that the CBN network is brittle to the \textit{type} and \textit{quality} of the contextual information and fails to generalize realistic scenarios by learning shortcuts.

\section{Background}
Since the introduction of Batch Normalization\cite{ioffe2015batch}, they have become an integral part of CNNs for faster convergence. The intermediary convolutional layers stabilise the variation of activations by normalising the input, $x_i$ characterised by the following equation:

\begin{equation}
    x_{i}^{N} = \frac{x_i - \mu_i}{\sigma_i}
\end{equation}

Batch normalization introduces learnable parameters, $\beta_i$ and $\gamma_i$ for each normalised layer $L_{i}^{N}$ : 
\begin{equation}
    x_{i}^{N} = (\frac{x_i - \mu_i}{\sigma_i+\epsilon})\gamma_i + \beta_i
\end{equation}
Conditional batch normalization\cite{de2017modulating}, aims to learn the learnable parameters from meta-data to "ground" visual features. It directly predicts affine scaling parameter of the affine transformation from meta-data. A shallow multi-layer perceptron, we call it auxiliary network, allows the meta-data to predict the $\beta^\prime$ and $\gamma^\prime$ that shifts and scales the normalised batch features. Formally, for meta-data $a_i$ and auxiliary network for $\mathcal{H}$ and $\mathcal{O}$ respectively for affine parameter prediction, $\mathcal{H}(a_i) = \beta^\prime$ and $\mathcal{O}(a_i) = \gamma^\prime$. 
\begin{equation}
    \hat{\beta_i} = \beta_i + \mathcal{H}(a_i) 
\end{equation}

\begin{equation}
        \hat{\gamma_i} = \gamma_i + \mathcal{O}(a_i) 
\end{equation}
These updated $\hat{\beta_i}$ and $\hat{\gamma_i}$ are used as parameters for batch normalization. 
Modulation of a neural network activations using conditional information through CBN has been proven successful in multi-modal setting \cite{perez2018film}. While CBN is a computationally efficient method to fuse different modalities, the robustness of this affine layer is unexplored. The addition of $\beta^\prime$ and $\gamma^\prime$, might seemingly be small yet showing a boost in performance of the task, we observe that it degrades in the quality of visual features learnt.

\section{Related Works}
Batch normalization\cite{ioffe2015batch} layers were introduced to alleviate the problem of internal covariate shift, also leading to faster model convergence. However, \cite{santurkar2018does} have argued that BN might not alleviate the internal covariate shift. Recent studies \cite{chang2015batch, yazdanpanah2022revisiting} have shown standard BN training is detrimental to performance in domain shift task. CBN layer was introduced to provide contextual information through affine transformation. \cite{perez2018film} uses CBN layer to modulate visual features to answer compositional questions on CLEVR dataset\cite{johnson2017clevr}. Conditional Instance Normalization \cite{dumoulin2016learned} was introduced in the context of style transfer. In reinforcement learning, \cite{DBLP:journals/corr/abs-1806-01946} used CBN layer to condition NMN and LSTM to follow compositionality in a 2D grid world. In generative modeling, \cite{brock2018large,zhang2019self} uses condition biasing through image description. \cite{oreshkin2018tadam} used CBN layer to make input distribution features robust to variations in few shot setting. While CBN has proved it efficacy through its acceptance in different applications but a thorough analysis of its potency is yet to be learnt\cite{michalski2019empirical}. We make the first attempt to uncover the potential risk of using CBN via a series of experiments.

\section{Experiments}
CBN is widely used for various applications such as multi-modal language acoustic learning\cite{abdelnour2018clear}, meta-learning\cite{requeima2019fast, zintgraf2019fast} and image generation\cite{brock2018large}. Due to its varied applications, we design our methodology for a supervised image classification task with visual and auxiliary features allowing for an easier surgery of the neural network. The aim of these experiments is to expose the potential pitfalls and impacts of fusing meta-data with images using CBN. 

\subsection{Experimental Setup}
We design our experiments to answer some fundamental questions posed by fusing of differently structured auxiliary data via CBN. We design a small experiment for each question posed although the major question, we aim to ask is - \textit{Does the addition of auxiliary data improve visual feature representation?}

\paragraph{CBN training - Baseline} In this setting, auxiliary data is passed as the input to the shallow auxiliary network that predicts the affine CBN parameters. This baseline experiment is mainly performed to establish effectiveness of CBN for classification. 

\paragraph{Effect on CBN in absence of test-time auxiliary data} 
The additional features passing through auxiliary network and its fusion with spatial features\cite{perez2018film} has been reported to improve accuracy across tasks. We aim to look at the neural network beyond final predictive accuracy with this experiment. In medical applications, the meta-data collection differs across hospitals, and often among different medical practitioners. We simulate this scenario by masking the attributes at test-time. Due to masking, the auxiliary network is switched off and the network only relies on its own affine transformation parameters. By reducing the dependence of network on auxiliary data, it is easier to establish its independence from visual features during test-time. We eventually completely switch off the auxiliary data stream and notice the difference in reliance of this auxiliary stream versus the spatial stream. 

\paragraph{Effect on CBN with masked train-time auxiliary data} 
While in previous experiment, we attempted to establish the independence of the visual stream, in this experiment, we aim to understand the quality of visual features learnt in absence of meta-data. We mask the attributes randomly to aid the network to learn and utilize visual features more. During masking, we essentially pass a flag that prevents the addition of $\mathcal{H}(a_i)$ and $\mathcal{O}(a_i)$ to $\hat{\beta_i}$ and $\hat{\gamma_i}$ respectively. 

\paragraph{Introduction of Supplementary network} We hypothesize that the presence of auxiliary data challenges the learning of visual features. Therefore we introduce a supplementary network to \textit{ground} and assist with the optimization of CBN network to learn visual features. The supplementary network has same network architecture as the CBN network but uses standard batch normalization. We optimize a loss consists of: cross-entropy loss of the CBN network, and distance between CBN network and supplementary network distribution. More formally, let $\mathcal{F}$ represent the CBN network that maps input $x_i$ to its label $y_i$ using subsidiary features, $a_i$ and $\mathcal{G}$ represent the BN network that maps input $x_i$ to its label $y_i$. Let the cross entropy loss of CBN network be represented by $\mathcal{L}^{CBN}$. We define the total sum as follows:

\begin{equation}
    Loss =  m * \mathcal{L}^{CBN}(x_i, y_i)  + (1-m) * D_{kl}(\mathcal{F}||\mathcal{G})
\end{equation}

where $m$ is a hyperparameter. It is important to note that the aim of supplementary network is to simply assist in optimization and grounding the CBN network, therefore the weighting over the distance between supplementary and CBN network is very small.

\paragraph{Removing data from the CBN network} Datasets such as CUB-200-211\cite{wah2011caltech} and Animals with Attributes\cite{xian2018zero} 2, which are popular in few-shot classification problem, have attributes which are highly correlated with the images and label. 
%We can easily learn about the effect of attributes on the CBN from previous experiments, in this experiment we aim to learn the impact of visual features on the CBN in isolation to auxiliary data.
We use these datasets to better study the learning capability of visual features in isolation from auxiliary data.

\subsection{Dataset}
We use natural images and medical images to test the effectiveness of CBN using our proposed training loss. We use \cite{wah2011caltech} dataset for the task of bird identification. It contains 11,788 images taken from 200 different species of bird. Every image of the dataset contains 312 binary and continuous concept labels. This dataset has been widely used in Few-Shot Classification and Conditional Image Generation, making it suitable for our analysis. The auxiliary data comes from the attributes. In our experiments we use binary attributes.

For medical images, Whole Slide Pathology images were used. The tumor-infiltrating lymphocytes (TILs)~\cite{saltz2018spatial} contains histological mappings based on each type of cancer. We use all 13 subsets of TCGA dataset, therefore constituting 13 cancer types. Although the popular task for such a dataset is necrosis classification~\cite{doi:10.1200/PO.18.00043} but since it is a binary classification task and is prone to overfitting without truly emphasising the effectiveness of our analysis, we modify the task to be a classification of different (here, thirteen) tumor types. The advantage of medical images is that their meta-data is readily available from diagnosis. A doctor uses whole slide images along with the meta-data for diagnostics indicating importance of fusion of different modalities. The meta-data includes information such as origin of tumor, age, gender, and size of tumor cells. 

\subsection{Results and Analysis}
In this section we evaluate the performance of networks that perform affine transformations using CBN and standard Batch Normalization. Due to BN's sensitivity to the batch size~\cite{lian2019revisit}, we perform all our experiments on batch size of 64. We perform our experiments on ResNet-18~\cite{he2015deep} backbone. For CBN network, we modify the ResNet-18 architecture to replace batch normalization layers with CBN layers. The auxiliary network that uses auxiliary data is a two layer MLP network to predict the affine parameters for CBN networks. The experiments in this paper were carried out using the V100SXM2 GPU. The results reported are averaged over 10 runs. Table \ref{tab:my_label} summarises the results on test set and we breakdown the analysis according to the set of different experiments below.

\begin{figure}[h] 
    \centering
    \includegraphics[width=1.0 \linewidth]{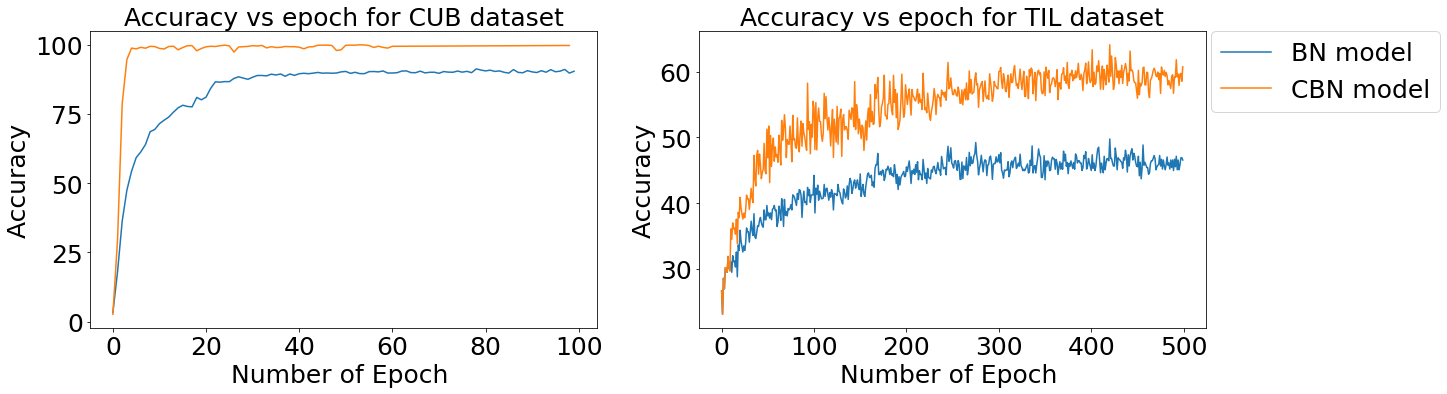}
    \caption{Validation accuracy versus epoch plotted for CUB dataset and TIL dataset respectively. Models with CBN layer perform better}
    \label{fig1}
\end{figure}

\paragraph{CBN Baseline} We train both CBN network and BN network to perform classification on the CUB and TIL dataset. In principle we expect the CBN network to learn from the spatial feature along with the features learnt from the other auxiliary modality. This suggests that, ideally, the representative power of this baseline model is greater than the representation from the standard vision only model. One may suspect that this should in turn lead to better generalization. From, Figure\ref{fig1} we observe that the validation accuracy of the CBN model is greater than the BN model by a great margin of $6.9\%$ in CUB and over $10\%$ in TIL dataset. This proves the effectiveness of modulating visual features by the auxiliary data. We interestingly also notice that the CBN network achieves convergence much faster than the BN network. Although convergence is a desired property, it is worrisome if the network converges in just 4 epochs achieving $100\%$ accuracy in the case of CUB dataset! Although, the same is not observed in TIL dataset, suggesting the difference in dataset and/or auxiliary data type. This particularly motivates the upcoming experiments to make an attempt at uncovering the black-box of conditional learning from auxiliary information stream. 

\begin{figure}[h] 
    \centering
    \includegraphics[width=0.8 \textwidth]{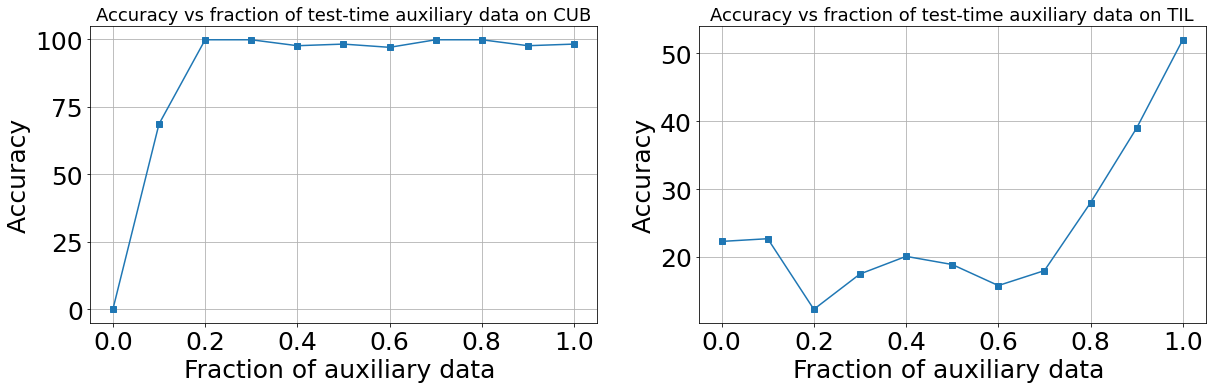}
    \caption{Accuracy versus the fraction of auxiliary data masked during inference.}
    \label{tt}
\end{figure}

\paragraph{Dependence of CBN on test-time auxiliary data} In a realistic scenario, due to data privacy and differences in data acquisition, meta-data collected for each patient maybe varied. If we were to leverage the CBN model in such a setting, it is likely that some of the auxiliary information is missing. To test this we limit the auxiliary information during validation and test-time in increments from 0 to 1 (full CBN network) of $10\%$ supplementary data. From Figure\ref{tt} we observe that in the case of CUB, with less than $20\%$ of auxiliary data, the network performs as well as in the presence of all of the auxiliary data. While in the case of TIL dataset, there is a sharp increase in the accuracy with greater fraction of meta-data. We hypothesize that this is due to the difference in structure and diversity of meta-data. Naturally histopathology data is very diverse suggestive of its difficulty in learning. In addition it must be noted that in complete absence of auxiliary information during testing time, the CBN network has very low accuracy, $0.5\%$ in the case of CUB and $22.3\%$ for TIL. This is suggestive of the network \emph{not learning visual} feature in the case of CUB dataset. While, simultaneously the CBN network does achieve greater than random accuracy in TIL dataset, its performance deteriorates highly in absence of the meta-data. 

\begin{figure}[h] 
    \centering
    \includegraphics[width=0.8 \textwidth]{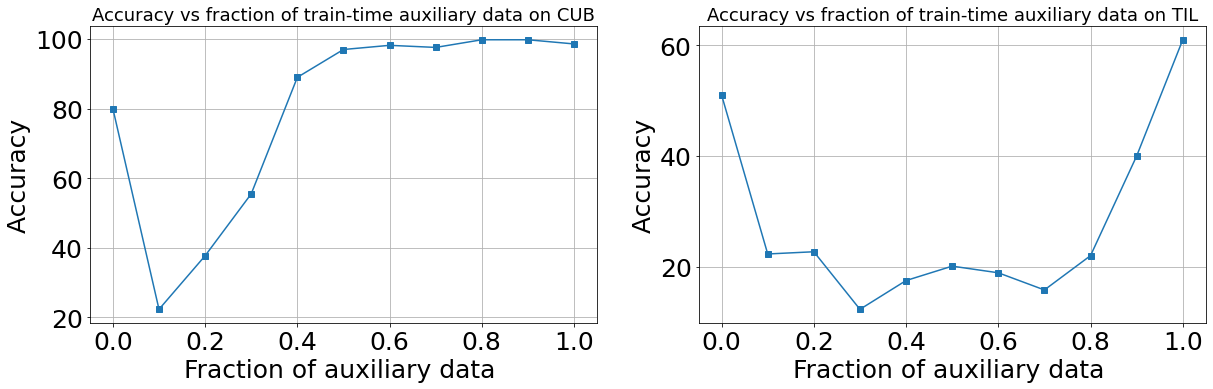}
    \caption{Accuracy versus the fraction of auxiliary data masked during training.}
    \label{traintime}
\end{figure}

\paragraph{Masking of train-time auxiliary data} 
From Figure~\ref{traintime}, the dependence of accuracy on the auxiliary features is apparent. We therefore attempt to establish the relationship between model's performance and the auxiliary data during training phase. In this experiment, we incrementally increase the fraction of auxiliary data fed in to the CBN model. With no auxiliary data at train-time, the network behaves similar to a BN model since auxiliary affine predictions are turned off, where as with all of the meta-data at train-time, the network behaves and performs similar to the baseline CBN model. In Figure~\ref{traintime}, we observe the general trend of achieving higher accuracy in presence of higher auxiliary data, although in the experiment with CUB dataset, the model converges much earlier with TIL data. Intuitively Figure~\ref{traintime} also signifies the inverse change in dependency of visual feature representation with increase in auxiliary data. It is observed that, with affine parameter prediction switches on, the neural network attempts to use the auxiliary data for label prediction.

\paragraph{Introduction of supplementary network} We hypothesized that supplementary network may help to ground and bring the CBN network closer to BN network, implying learning of visual features. Hyperparamaters $m$ equals $0.9$. We observe that the hypothesis is proven incorrect and supplementary network is not able to reduce the dependence of auxiliary data on CBN network. In Figure\ref{appendix} we observe that the accuracy curves for both CUB and TIL datasets are very similar to CBN curve, although more stable. Due to ineffectiveness of the supplementary network on baseline experiment, we perform all our experiments with CBN network.

\paragraph{Removing spatial data} Table~\ref{tab:my_label} present that even in the absence of image data, the network is able to perform as well as the baseline CBN network for CUB dataset. This is highly suggestive of presence of shortcuts in data and the visual features not contributing to the CBN model's predictive performance. On the contrary, CBN network on the TIL dataset learns some feature in absence of visual features, therefore indicating partial learning of vision features. 

\begin{figure}[]
     \centering
     \begin{subfigure}[b]{0.33\textwidth}
         \centering
         \caption{CBN model}
         \includegraphics[width=1\textwidth]{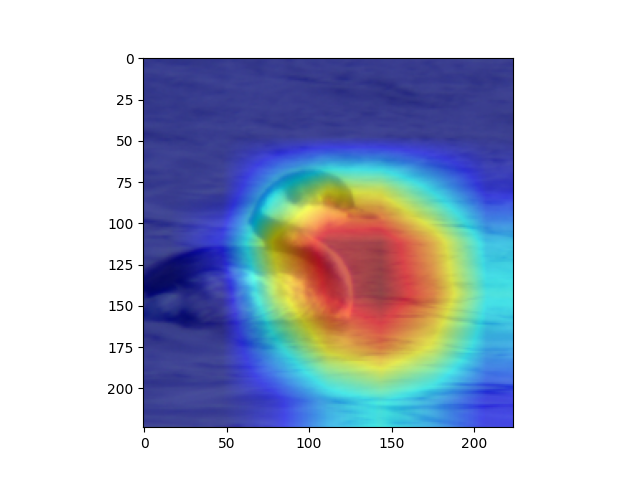}
         
         \label{fig:til_cms}
     \end{subfigure}
     \hfill
     \begin{subfigure}[b]{0.33\textwidth}
         \centering
         \caption{CBN with 0 attributes at test-time}
         \includegraphics[width=\textwidth]{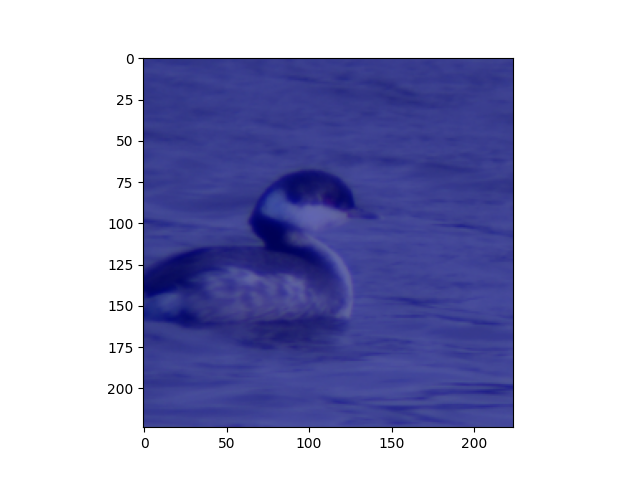}
         
         \label{fig:til_cmt}
     \end{subfigure}
     \begin{subfigure}[b]{0.32\textwidth}
         \centering
         \caption{BN model}
         \includegraphics[width=\textwidth]{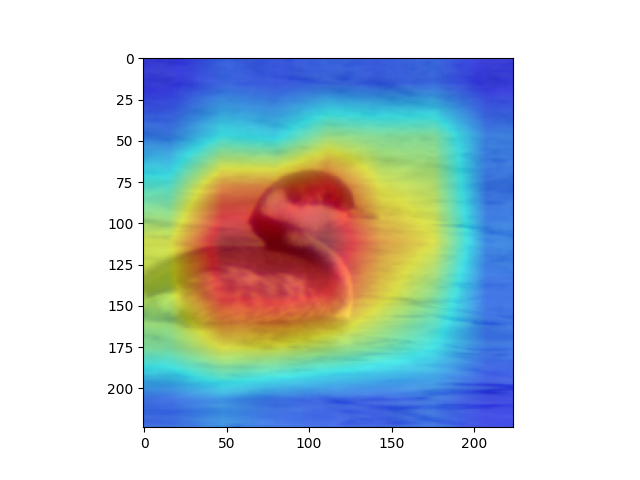}
         
         \label{fig:til_cm}
     \end{subfigure}
        \caption{A visualization of GradCAM\cite{selvaraju2017grad} features for \textit{Horned Grebe} class. Although the CBN model achieve $100\%$ accuracy, the feature of BN model are visually more descriptive.}
        \label{fig:bird_cam}
\end{figure}

\paragraph{Discussion} From an exhaustive set of experiments, we observe that although the attenuating the affine batch normalization parameter via a second modality leads is higher predictive performance as compared to standard batch normalization training of vision modality, it leads to deterioration of visual features. In the case of both TIL and CUB, there is an increased dependency on the auxiliary data. This suggests that the affine parameters of CBN network allow for a shortcut learning of auxiliary data leading to poor generalization. Visually we compare the features learnt by CBN network with all attributes, CBN network with no attributes and BN network for both CUB and TIL dataset. We observe that the features learnt in each of the cases are very different signifying the potential difference in their respective minima. It is particularly interesting to note that in the absence of visual features in CUB, the network has random predictive performance signified by \ref{fig:til_cmt}. We observe that the visual features presented in \ref{fig:til_cms} is less explanatory of the bird features as compared to \ref{fig:til_cmt}.

\begin{figure}[h]
     \centering
     \begin{subfigure}[b]{0.33\textwidth}
         \centering
         \caption{CBN model}
         \includegraphics[width=1\textwidth]{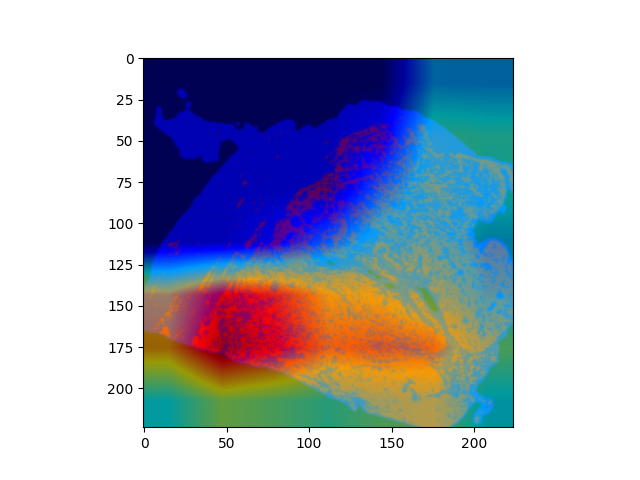}
         
         \label{fig:cms}
     \end{subfigure}
     \hfill
     \begin{subfigure}[b]{0.33\textwidth}
         \centering
         \caption{CBN with 0 attributes at test-time}
         \includegraphics[width=\textwidth]{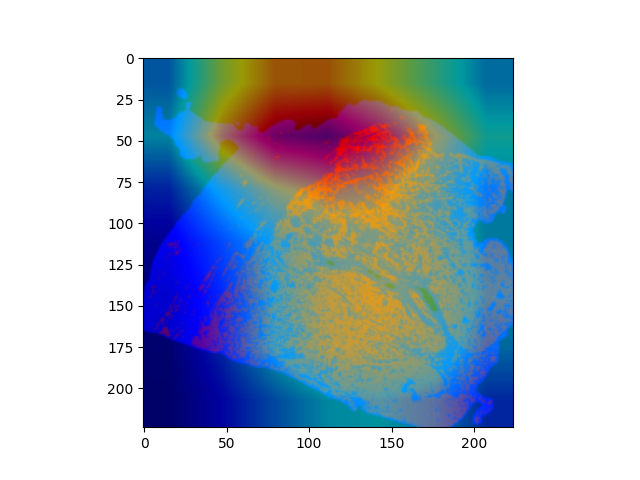}
         
         \label{fig:cmt}
     \end{subfigure}
     \begin{subfigure}[b]{0.32\textwidth}
         \centering
         \caption{BN model}
         \includegraphics[width=\textwidth]{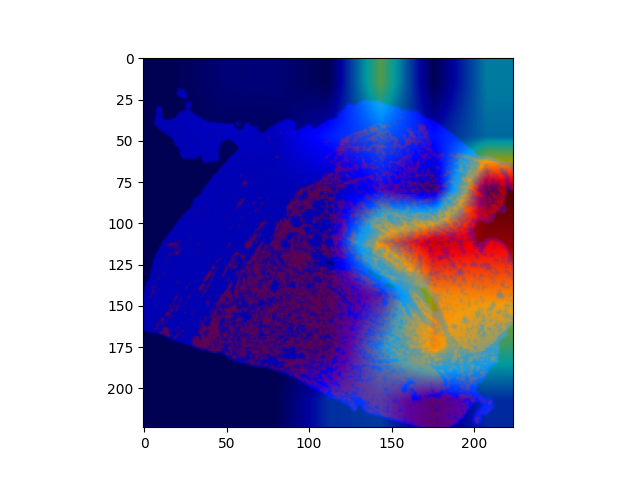}
         
         \label{fig:cmt}
     \end{subfigure}
        \caption{A visualization of GradCAM\cite{selvaraju2017grad} features for \textit{BLCA} cancer type class.}
        \label{fig:til_cam}
\end{figure}

\paragraph{Ablation study - Continuous vs Discrete attributes in CUB}
The CUB dataset contains continuous attribute values between 0 and 1. We threshold the values greater than $50$ to $1$ and rest to $0$. Although in the case of continuous attributes, we observe from Figure~\ref{attri} that the neural network converges within 2 epochs which suggests that the continuous attributes leads to memorization, highly undesirable for generalization. Therefore we use binary attributes in all of our experiments.

%%%%%%%%%%%%%%%%%%%%%%%%%%%%%%%%%%%%%%%%%%%%%%%%%%%%%%%%%%%%
\section{Conclusion}
In this work, we explored the efficacy of Conditional Batch Normalization to fuse an auxiliary network with a vision only model. We observe that the CBN network motivates learning of highly undesirable shortcuts learning between the auxiliary data and the label. This leads to a boost in performance but a drop in visual features. We observed that the effectiveness of CBN is dependent on the dataset. We hope that this work inspires research in the direction of learning generalizable visual features while conditioning in presence of differently structured modality. 
% \appendix

\section{Acknowledgements}
Authors would like to thank Vincent Michalski for fruitful discussions. We would like to thank the Digital Research Alliance of Canada for compute resources, Google and CIFAR for research funding. IS acknowledges funding from Mitacs and Imagia Canexia Health. AAR acknowledges funding from FRQS. 

\bibliographystyle{plain}
\bibliography{ref}
% \bibliography{ref}
%\printbibliography

\appendix
\section*{Appendix}

\begin{table}[H]
    \centering
    \begin{tabular}{ll|r|r}
    \toprule
        Training Model & Testing Model & CUB  & TIL \\ \midrule
        
        BN & BN &$\underset{\pm 1.1}{79.9}$ &$\underset{\pm 2.3}{51.1}$ \\ \midrule
        CBN Baseline & CBN Baseline &$\underset{\pm 0.6}{99.2}$ &$\underset{\pm 3.4}{62.9}$ \\ \midrule
        CBN Baseline & CBN-No attributes &$\underset{\pm 0.1}{0.5}$ &$\underset{\pm 3.8}{19.3}$ \\ \midrule
        CBN + Supplementary Baseline& CBN + Supplementary Baseline &$\underset{\pm 0.5}{99.0}$ &$\underset{\pm 2.9}{63.1}$ \\ \midrule
        CBN-No visual data & CBN-No visual data &$\underset{\pm 0.6}{99.1}$ &$\underset{\pm 10.7}{20.8}$ \\ \midrule
    \end{tabular}
    \caption{Summary of results}
    \label{tab:my_label}
\end{table}

\begin{figure}[h] 
    \centering
    \includegraphics[width=1.0 \textwidth]{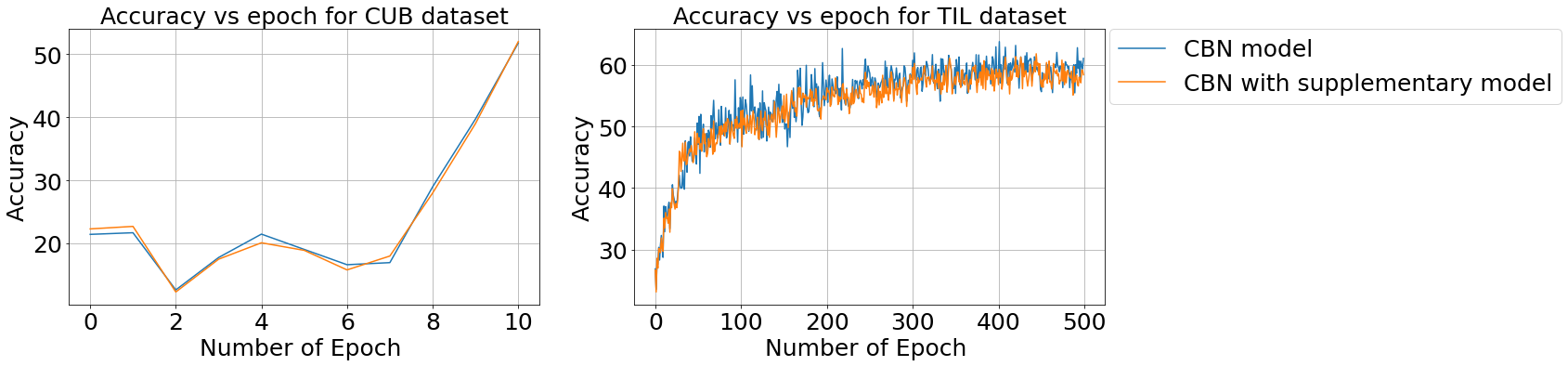}
    \caption{Accuracy versus epoch for CBN hybrid in presence of supplementary network.}
    \label{appendix}
\end{figure}

\begin{figure}[h] 
    \centering
    \includegraphics[width=0.6 \textwidth]{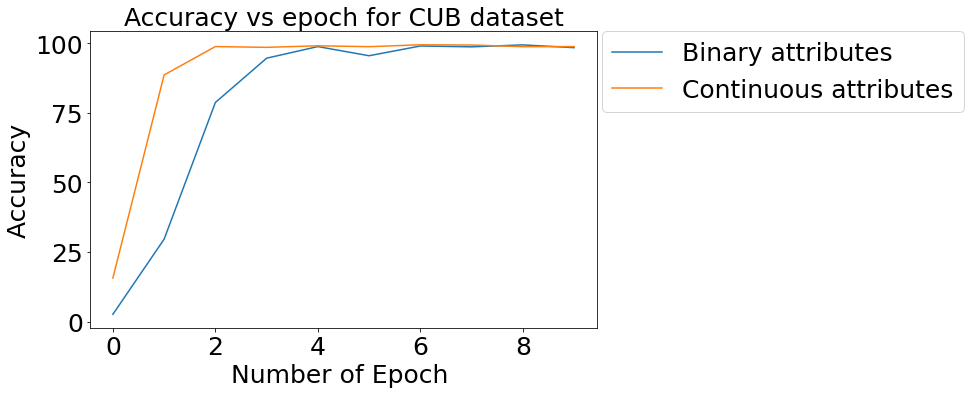}
    \caption{Accuracy versus epoch for CBN network in presence of continuous and binary attributes. }
    \label{attri}
\end{figure}

\end{document}